\pdfoutput=1

\documentclass[11pt]{article}

\usepackage{amsmath}
\usepackage{graphicx} 

\usepackage[]{ACL2023}

\usepackage{times}
\usepackage{latexsym}
\usepackage{booktabs}
\usepackage{multirow}

\usepackage[T1]{fontenc}

\usepackage[utf8]{inputenc}

\usepackage{microtype}

\usepackage{inconsolata}

\title{Leveraging Entity Information for Cross-Modality Correlation Learning: The Entity-Guided Multimodal Summarization}

\author{Yanghai Zhang,  Ye Liu,  Shiwei Wu,  Kai Zhang\thanks{\ \ Corresponding author.}, \\
{\bf Xukai Liu,  Qi Liu,  Enhong Chen}\\
State Key Laboratory of Cognitive Intelligence, \\
University of Science and Technology of China, Hefei, China \\
\texttt{\{yhzhang0612, liuyer, dwustc, chthollylxk\}@mail.ustc.edu.cn} \\ \texttt{\{kkzhang08, qiliuql, cheneh\}@ustc.edu.cn}
}

\begin{document}
\maketitle
\begin{abstract}
The rapid increase in multimedia data has spurred advancements in Multimodal Summarization with Multimodal Output (MSMO), which aims to produce a multimodal summary that integrates both text and relevant images. The inherent heterogeneity of content within multimodal inputs and outputs presents a significant challenge to the execution of MSMO. Traditional approaches typically adopt a holistic perspective on coarse image-text data or individual visual objects, overlooking the essential connections between objects and the entities they represent. To integrate the fine-grained entity knowledge, we propose an Entity-Guided Multimodal Summarization model (EGMS). Our model, building on BART, utilizes dual multimodal encoders with shared weights to process text-image and entity-image information concurrently. A gating mechanism then combines visual data for enhanced textual summary generation, while image selection is refined through knowledge distillation from a pre-trained vision-language model. Extensive experiments on public MSMO dataset validate the superiority of the EGMS method, which also prove the necessity to incorporate entity information into MSMO problem.
\end{abstract}

\section{Introduction}
\begin{figure}[t]
\centering
\includegraphics[width=1.0\columnwidth]{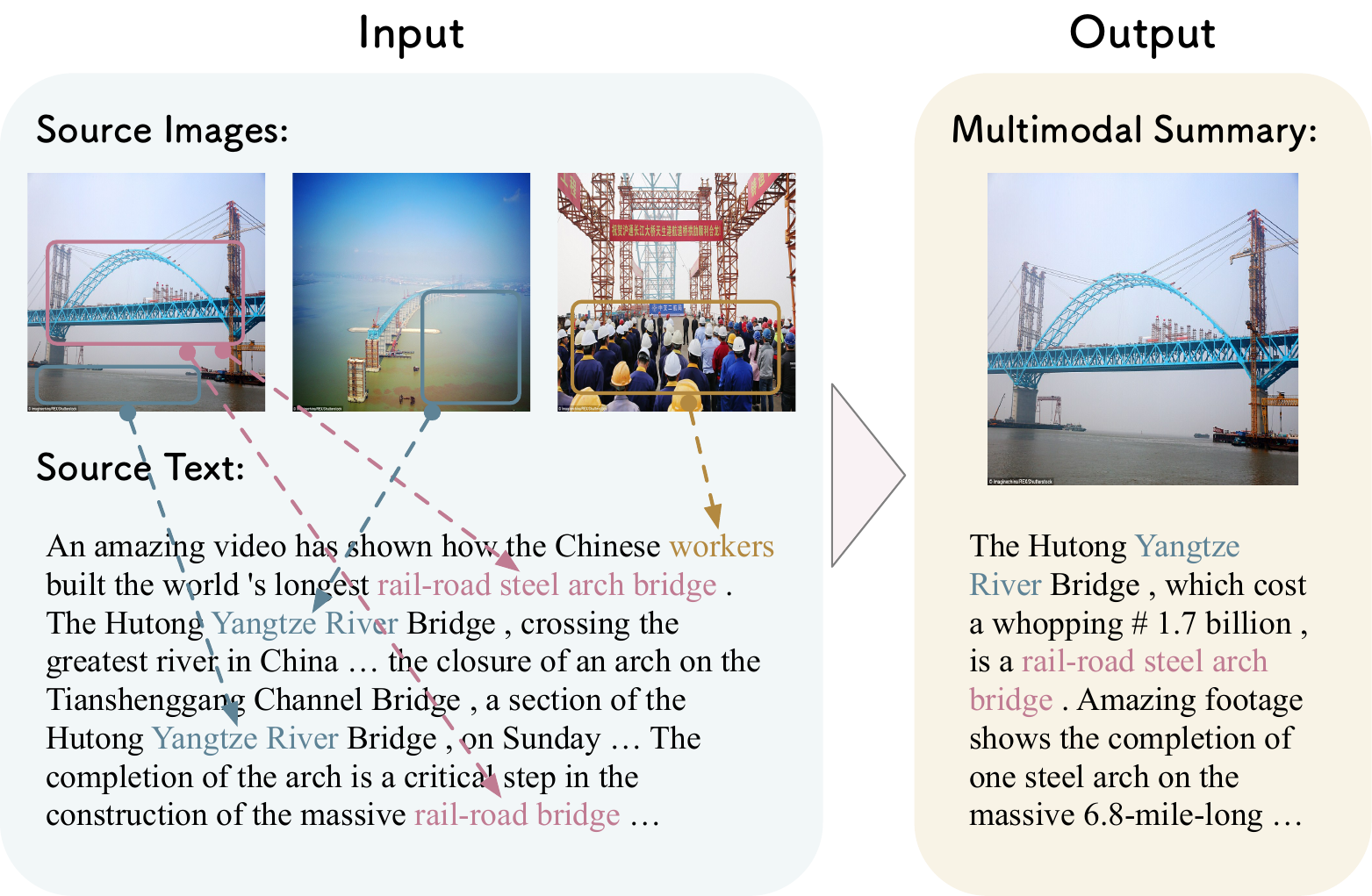}
\caption{An example of entity-object correlations in multimodal data from MSMO problem. 
Entities \textit{rail-road steel arch bridge} and \textit{Yangtze River} correspond with elements in the associated images, suggesting inherent cross-modality correlations.
}
\label{fig:case}
\end{figure}

With the rapid development of multimedia content across the Internet, the task of Multimodal Summarization with Multimodal Output (MSMO) has emerged as a research direction of considerable significance \citep{zhu-etal-2018-msmo, zhu2020multimodal, mukherjee-etal-2022-topic, zhang2022unims, zhang2022hierarchical}, especially for news content summary \citep{zhu-etal-2018-msmo}. Specifically, as shown in Figure~\ref{fig:case}, given the source text and corresponding images, MSMO aims to produce a multimodal summary with a textual abstract alongside a pertinent image. Instead of providing exclusively text-based summaries, MSMO considers and generates more diverse multimodal information, which constitutes a significant research but also puts high challenges for the interaction between text and images \citep{zhu2020multimodal}.

Since \citet{zhu-etal-2018-msmo} proposed the MSMO task and collected the first large-scale English corpus, there has been a surge of research in academia exploring this area.
However, most existing methodologies \citep{zhu-etal-2018-msmo, zhu2020multimodal, mukherjee-etal-2022-topic, zhang2022unims} integrated comprehensive image and text data without allocating explicit attention to discrete constituents within these modalities. \citet{zhang2022hierarchical} have made strides in enhancing the domain by facilitating interactions between textual components at the granular word level and discrete objects in visual content. Nonetheless, these visual objects tend to relate to entity-level content in text rather than individual words. For example, from Figure \ref{fig:case}, we can see that multi-word entities \textit{rail-road steel arch bridge} and \textit{Yangtze River} within the textual corpus exhibit correspondence with elements depicted in the accompanying images. Similar to how \citet{zhang2019interactive} accounted for aspect granularity and \citet{liu2023techpat} utilized phrase-level information in text analysis, we aim to leverage entity information. Explicit extraction of these entities is posited to enhance comprehension of the image content. To our knowledge, few works have focused on incorporating entity information into MSMO problem.

Indeed, there are many technical challenges inherent in designing effective solutions to incorporate entity information into MSMO process.
The first of these pertains to the heterogeneity of the data involved, which can be textual, pictorial, or entity-based. This diversity imposes significant hurdles in attaining efficient cross-modality interaction.
Second, the traditional frameworks employed for text decoding are predominantly designed to process purely textual inputs, thus creating a conundrum when the need arises to incorporating multi-modal data into the decoding procedure.
Third, the task of image selection, which tends to operate independently, frequently suffers from an absence of adequate labeling information, as there are no golden labels in the training set. 

To address these challenges, we propose an Entity-Guided Multimodal Summarization model (EGMS). Building on the BART framework, similar to UniMS \citep{zhang2022unims}, we reconfigure BART's text-centric encoder to create a Shared Multimodal Encoder. This encoder includes two multimodal encoders with shared parameter weights to model textual and visual data alongside entity-specific visual information. For the decoding process, we design a Multimodal Guided Text Decoder. It employs a gated image fusion module to effectively merge image representations enriched with multimodal information and utilizes this information for text generation. Subsequently, we introduce a Gated Knowledge Distillation module to leverage the expertise of a pre-trained vision-language model, which serves as an auxiliary guide for the image selection learning process.
Finally, we conduct extensive experiments on public MSMO datasets, where the experimental results demonstrate the effectiveness of our proposed EGMS method.
Our code is available via \url{https://github.com/ApocalypseH/EGMS}.

\section{Related Work}

\subsection{Multimodal Summarization}
Multimodal summarization \citep{uzzaman2011multimodal} is defined as a task that aims at distilling concise and precise syntheses from heterogeneous data sources, encompassing textual, visual, and audio content, etc. 
Research endeavors \citep{li-etal-2017-multi, chen-zhuge-2018-abstractive, li2018multi, zhang2021ctnr} have predominantly concentrated on the incorporation of supplementary and ancillary modal information to augment the depiction of a solitary modality. 
For example, \citet{li2018multi} design image filters with the intent to selectively harness visual information, thereby augmenting the semantic richness of the input sentence.

Recently, there has been a burgeoning interest in the domain of multimodal summarization with multimodal output (MSMO). 
\citet{zhu-etal-2018-msmo} construct the first large-scale corpus MSMO for this novel summarization task, which integrates textual and visual inputs to produce a comprehensive pictorial summary. They also propose a multimodal attention framework to jointly synthesize textual summary and select the most relevant image. 
Then \citet{zhu2020multimodal} introduce a novel evaluation metric that integrates multimodal data to better combine visual and textual content during both the training and assessment stages.
\citet{mukherjee-etal-2022-topic} and \citet{zhang2022unims} propose to solve the multimodal summarization task in a multitask training manner.
And \citet{zhang2022hierarchical} adopt a graph network and a hierarchical fusion framework to learn the intra-modal and inter-modal correlations inherent in the multimodal data respectively.

\subsection{Knowledge Graph Augmented Models}
Knowledge Graphs (KGs) store and organize information about different things and how they relate to each other in a structual way. World knowledge is commonly expressed using fact triplets, which consist of three elements: the subject entity, the relation, and the object entity denoted as $(h, r, t)$. Since  the introduction of TransE \citep{bordes2013translating}, a multitude of knowledge graph embedding techniques \citep{ji2015knowledge, zhong2015aligning, shi2017proje} have emerged, aiming to translate the entities and relationships within these graphs into numerical vectors so that they can be easily applied to various downstream tasks.

\citet{zhang-etal-2019-integrating} and \citet{liu-etal-2023-enhancing} leverage external knowledge graphs to enhance the textual content for improved performance in text classification tasks.
Moreover, \citet{hu-etal-2022-knowledgeable} concentrate on the integration of external knowledge into the verbalizer mechanism to enhance the effectiveness and stability of prompt tuning for zero and few-shot text classification tasks.
\citet{yu-etal-2022-kg} improve Fusion-in-Decoder \citep{izacard-grave-2021-leveraging} by employing a knowledge graph to establish the structural interconnections among the retrieved passages in Open-Domain Question Answering (ODQA) problem, achieving comparable or better performance with a much lower computation cost.

\section{Preliminary} 

\subsection{Problem Formulation}
Given a multimodal input $D = \{T, P\}$, where $T = \{t_1, t_2, ..., t_L\}$ is a sequence of $L$ tokens of the article text and $P = \{p_1, p_2, ..., p_M\}$ is the collection of the $M$ in-article images, our proposed model first extracts all the entities $K = \{k_1, k_2, ..., k_N\}$ in the article text and then summarizes $\{D, K\}$ into a multimodal summary $S = \{S_t, S_i\}$. $S_t = \{s_1, s_2, ..., s_l\}$ denotes the textual summary limited by a max length of $l$. The pictorial summary $S_i$ is an extracted subset of the image input $P$.

\subsection{BART Architecture}
BART (Bidirectional and Auto-Regressive Transformers) \citep{lewis-etal-2020-bart} functions as a denoising autoencoder, designed to reconstruct an original document from its corrupted counterpart. 
\begin{figure}[h]
\centering
\includegraphics[width=1.0\columnwidth]{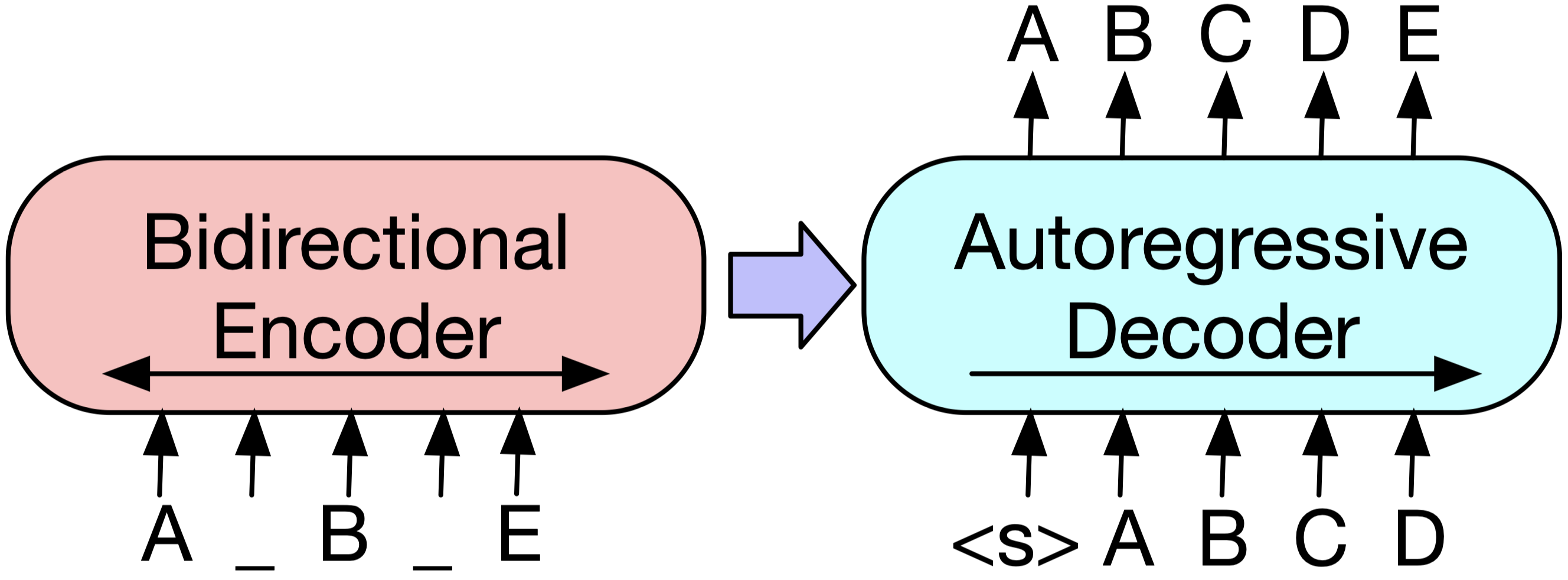}
\caption{BART architecture from \citet{lewis-etal-2020-bart}.}
\label{fig:bart}
\end{figure}

As shown in Figure \ref{fig:bart}, it uses a standard Transformer-based neural machine translation architecture, incorporating a bidirectional encoder, coupled with a left-to-right autoregressive decoder.
In the process of optimizing BART for text generation applications, the source text is initially fed into the encoder module. Following this, the desired output text, which is prepended with the decoder's designated initial token, is introduced to the decoder module.

\section{Model}
\label{sec:model}
\begin{figure*}[!t]
\centering
\includegraphics[width=2.0\columnwidth]{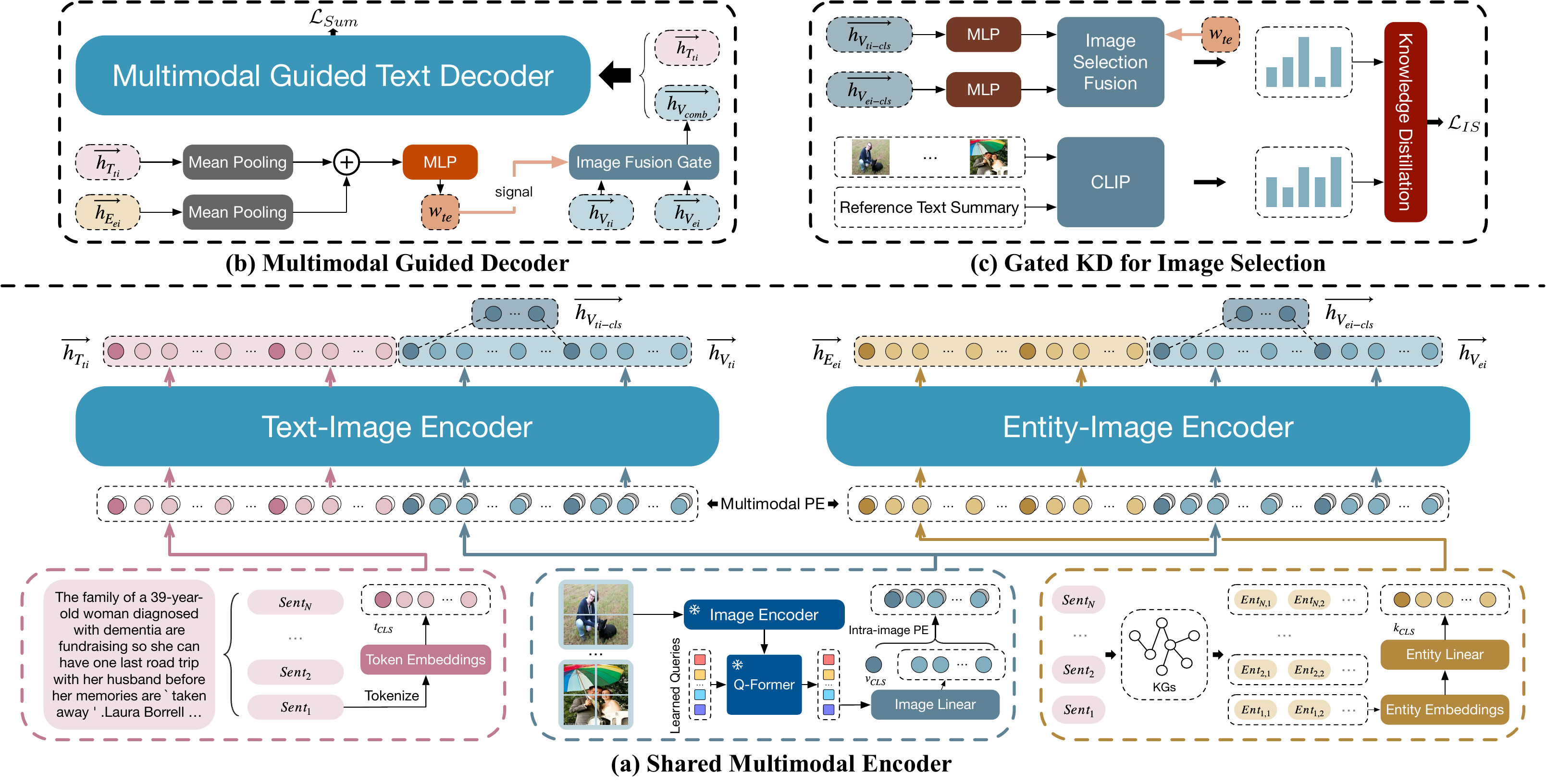}
\caption{The architecture of our proposed EGMS model. It consists of three parts: 
(a) Shared Multimodal Encoder; (b) Multimodal Guided Decoder; (c) Gated Knowledge Distillation for Image Selection.
}
\label{fig:model}
\end{figure*}
\subsection{Model Overview}
We propose a novel multimodal summarization framework enhanced by an external knowledge graph, as shown in Figure \ref{fig:model}. Building upon the BART architecture, our model has been adapted to accommodate multimodal inputs, specifically textual and visual data. Recognizing that images often depict objects which correspond to real-world entities, our approach seeks to leverage this multimodal data more effectively. To this end, we utilize an external knowledge graph to extract entities from the textual content, which in turn facilitates a better interpretation of the visual information. This integration aims to improve the coherence and richness of the generated summaries by bridging the semantic gap between the textual and visual modalities.

\subsection{Shared Multimodal Encoder}
\paragraph{Text-Image Encoder}
Given the inherent restriction of BART's context length, capped at 1024 tokens, it is imperative to deliberate on the regulation of image input dimensions to ensure compatibility with the model's processing capabilities. Following \citet{li2023blip}, we use a frozen Q-Former to transform image features $r_i^{|IE| \times d_{IE}}$, which are derived from a frozen image encoder, into a fixed number of output features $v_{i}^{|Q| \times d_Q}$, each corresponding to a predefined learned query $q$:
\begin{equation}
    \begin{gathered}
        r_i = [r_{i,1}, r_{i,2}, ..., r_{i,|IE|}] = f_{img-enc}(p_i), \\
        \begin{aligned}
            v_i &= [v_{i,1}, v_{i,2}, ..., v_{i,|Q|}] \\
                &= f_{Q-Former}(q_1, q_2, ..., q_{|Q|}; r_i),
        \end{aligned}
    \end{gathered}
\end{equation}

Then, we enhance the textual encoding capabilities of BART by transitioning to a multimodal encoding framework. For text-image encoder, this involves the integration of textual embeddings, denoted as $e_t$, with corresponding visual embeddings $e_v$. The concatenated embeddings serve as input to the encoder $f_{ti-enc}$, which then yields contextualized representations:
\begin{equation}
    \begin{gathered}
        e_t = W_{t} \cdot [t_{CLS}, t_1, t_2, ..., t_L, t_{SEP}] , \\
        e_{v_i} = [v_{CLS}, W_{v} \cdot v_{i}] + e_{intra-pos}, \\
        \begin{aligned}
            e_{ti} &= [e_t, e_v] + e_{multi-pos} \\
                   &= [e_t, e_{v_1}, ..., e_{v_M}] + e_{multi-pos},
        \end{aligned} \\
        h_{ti} = [h_{T_{ti}}, h_{V_{ti}}] = f_{ti-enc}(e_{ti}),
    \end{gathered}
\end{equation}
where special tokens $t_{CLS}$ and $t_{SEP}$ serve as delimiters to denote the start and end of each sentence respectively. The embeddings $e_{intra-pos}$ and $e_{multi-pos}$ represent the intra-image positional information and the multimodal positional context within the framework. The matrices $W_{t}$ and $W_{v}$ are employed for embedding linguistic tokens and projecting image features into a shared multimodal space respectively.  Following \citet{dosovitskiy2021an} and \citet{zhang2022unims}, we add a learnable special token, represented by the embedding vector $v_{CLS}$, to signify the initiation of an image sequence. The corresponding encoded state at the output of the encoder is then utilized as a holistic representation of the image.

\paragraph{Entity-Image Encoder}
For the reasons already explained in the introduction, we propose to incorporate entity-level information to enhance the exploitation of multimodal data.

First, we extract entities from the text utilizing an external knowledge graph. For the clarity and simplicity, we adopt the classical TransE model \citep{bordes2013translating} to obtain a representation of the entities in the knowledge graph, which contains intricate structural relationships among the entities. Similar to the text-image encoder, the entity embeddings $e_e$ concatenated with visual embeddings $e_v$ are subsequently processed by the entity-image encoder $f_{ei-enc}$, yielding an enriched image representation that encapsulates augmented entity-specific information:
\begin{equation}
    \begin{gathered}
        e_e =  W_{e_2} \cdot W_{e_1} \cdot [k_{CLS}, k_1, k_2, ..., k_M], \\
        e_{ei} = [e_e, e_v] + e_{multi-pos}, \\
        h_{ei} = [h_{E_{ei}}, h_{V_{ei}}] = f_{ei-enc}(e_{ei}),
    \end{gathered}
\end{equation}
where $k_{CLS}$ is used to demarcate sequences of entities contained in discrete sentences. The matrix $W_{e_1}$ represents the embedding matrix for entities, which is initialized utilizing embeddings derived from the pre-trained TransE model. Concurrently, the matrix $W_{e_2}$ is employed to project entity features into a unified multimodal space for further integration of modalities. Notably, this encoder shares its parameter weights with the aforementioned text-image encoder.

\subsection{Multimodal Guided Decoder}
\paragraph{Gated Image Fusion}

Inspired by \citet{zhang-etal-2022-incorporating}, we introduce a gated image fusion module to integrate the visual representations derived from dual encoders, each amalgamated with textual and entity-based information respectively. Visual information integrated with textual and entity representations from the respective encoders will be merged together:
\begin{equation}
    h_{te} = Mean(h_{T_{ti}}) \oplus Mean(h_{E_{ei}}),
\end{equation}
where $\oplus$ is the concatenation operation.

Then $h_{te}$ will serve as the input for a weight computation module, which is designed to quantitatively assess the salience of the visual representations in conjunction with corresponding multimodal inputs:
\begin{equation}
    \label{weight}
    w_{te} = \sigma_{w}^{2} (W_w^2 \cdot \sigma_{w}^{1}(W_w^1 \cdot h_{te} + b_w^1) + b_w^2),
\end{equation}
where $\sigma_{w}^{2}$ is $Sigmoid$ activation function, making the value of this weight between 0 and 1.

Subsequently, the derived weight $w_{te}$ serves as the signal to control the fusion of dual image representations that encapsulate different modal information, yielding an augmented image representation that is enriched with both textual and entity information:
\begin{equation}
    h_{V_{comb}} = w_{te} \cdot h_{V_{ti}} + (1 - w_{te}) \cdot h_{V_{ei}}.
\end{equation}

\paragraph{Multimodal Guided Text Decoder}
Similar to BART, the architecture of our model incorporates a conventional autoregressive transformer decoder within its decoding module. In contrast to relying exclusively on textual representations during the encoding phase, our proposed model also utilizes the aforementioned augmented image representations. These representations serve as encoder hidden states that are subsequently fed into the decoder:
\begin{equation}
    h_{enc-hid} = [h_{T_{ti}}, h_{V_{comb}}].
\end{equation}

The decoder attends to the sequence of previously generated tokens, denoted as $s_{<j}$, as well as the encoder output hidden states $h_{enc-hid}$, and predicts the conditional probability distribution of subsequent text tokens. So for the abstractive summarization task, our model is trained by minimizing the negative log-likelihood:
\begin{equation}
    \mathcal{L}_{Sum} = - \sum_{j=1}^{|S|} \log p(s_j|s_{<j},h_{enc-hid}, \phi),
\end{equation}
where $\phi$ denotes all the parameters of the model.

\subsection{Gated Knowledge Distillation for Image Selection}
In the current multimodal summarization dataset, only the test set has visual references, which could be instrumental in guiding the selection of salient images during the training phase. 


\citet{zhang2022unims} propose to adopt Knowledge Distillation (KD) technique \citep{hinton2015distilling} to distill the inherent relevance between textual and visual information, which can get image references without any image captions or visual references. Rather than using only the text-integrated image representations as \citet{zhang2022unims}, we incorporate entity information as well. Specifically, we use the output hidden states of $v_{CLS}$ derived from both encoders as comprehensive image representations and feed them to two distinct multi-layer perceptrons to obtain scores:
\begin{equation}
    \begin{gathered}
        g_{ti}(p) = W_{t}^2 \cdot \sigma_{t}^{1}(W_{t}^1 \cdot h_{v_{ti-cls}} + b_{t}^1) + b_{t}^2, \\
        g_{ei}(p) = W_{e}^2 \cdot \sigma_{e}^{1}(W_{e}^1 \cdot h_{v_{ei-cls}} + b_{e}^1) + b_{e}^2.
    \end{gathered}
\end{equation}

And we combine them with the weight calculated in Eq.(\ref{weight}) to futher utilize multimodal information:
\begin{equation}
\label{image-score}
    g(p) = w_{te} \cdot g_{ti}(p) + (1 - w_{te}) \cdot g_{ei}(p).
\end{equation}

We employ CLIP \citep{radford2021learning} as the teacher model to calculate the similarity scores between each image $p$ and the textual summary $S_t$:
\begin{equation}
    l(S_t, p) = sim(\mathcal{T}(S_t), \mathcal{V}(p)),
\end{equation}
where $\mathcal{T}$ and $\mathcal{V}$ are its textual and visual encoder respectively, and $sim$ is the cosine similarity function.

Through knowledge distillation, our model is intended to emulate the score distribution of the teacher model. By using Kullback-Leibler (KL) divergence \citep{kullback1951information}, this approach can be modeled as minimizing the following objective function with temperature $\tau$:
\begin{equation}
    \mathcal{P}_p(p, \tau)
    =
    \frac{\exp(\frac{g(p)}{\tau})}{\sum_{p \in P}\exp(\frac{g(p)}{\tau})},
\end{equation}

\begin{equation}
    \mathcal{Q}_p(S_t, p, \tau)
    =
    \frac{\exp(\frac{l(S_t, p)}{\tau})}{\sum_{p \in P}\exp(\frac{l(S_t, p)}{\tau})},
\end{equation}

\begin{equation}
    \mathcal{L}_{IS} = KL(\mathcal{P}||\mathcal{Q}) = - \sum_{p \in P}\mathcal{P}_p \cdot \ln \frac{\mathcal{Q}_p}{\mathcal{P}_p}.
\end{equation}

\subsection{Training}
Inspired by \citet{li2023blip}, we divide the training process of our proposed model into two main stages: an initial phase dedicated to aligning the modalities of images and text, followed by a subsequent phase focusing on fine-tuning.
\paragraph{Modal Matching}
In the modal matching phase, parameter optimization is confined to the weights of the image feature projection matrix $W_v$, and the embedding $v_{CLS}$ of the visual initiation token. This targeted approach leverages the text-image encoder and the decoder exclusively, thereby enhancing the model's focus on the pertinent multimodal information while alleviating the impact of other information. The training process is governed by minimizing the negative log-likelihood:
\begin{equation}
    \begin{aligned}
        \mathcal{L} &=  - \sum_{j=1}^{|S|} \log p(s_j|s_{<j},h_{ti}, \phi), \\
        &=  - \sum_{j=1}^{|S|} \log p(s_j|s_{<j},[h_{T_{ti}}, h_{V_{ti}}], \phi).
    \end{aligned}
\end{equation}

\paragraph{Fine-tuning}
In the fine-tuning phase, the model parameters are initially set using the weights obtained from the modal matching stage. Subsequently, the entire proposed framework is employed, with adjustments made to all learnable parameter weights. Similar to \citet{zhang2021eatn}, the training loss of our model comprises the objectives of its various components, specifically image selection and abstractive text summarization:
\begin{equation}
    \mathcal{L} = \alpha \cdot \mathcal{L}_{IS} + \mathcal{L}_{Sum},
\end{equation}
where $\alpha$ is a hyper-parameter that modulates the salience of the image selection loss within the total training loss.

\section{Experiment}
\subsection{Experiment Setup}
\paragraph{Datasets}
For multimodal summarization with multimodal output, we use the MSMO dataset, which is introduced by \citet{zhu-etal-2018-msmo}. This is the first and only large-scale English corpus specifically curated for this task.
It comprises a collection of online news articles sourced from $Daily Mail$ website\footnote{\url{http://www.dailymail.co.uk}}, each accompanied by several images and corresponding manually-written highlights that serve as the reference summary. More statistics about the dataset are illustrated in Table \ref{tab:msmo}. Within the test set, a maximum of three images are annotated to provide a pictorial reference.
\begin{table}[]
\centering
\begin{tabular}{lccc}
\toprule
\textbf{Statistics} & \textbf{Train} & \textbf{Valid} & \textbf{Test}\\
\midrule
\#Samples & 293,965 & 10,355 & 10,261 \\
\#AvgTokens(A) & 720.87 & 766.08 & 730.80 \\
\#AvgTokens(S) & 70.12 & 70.02 & 72.16 \\ 
\#AvgImgs & 6.56 & 6.62 & 6.97  \\
\bottomrule
\end{tabular}
\caption{The data statistics of MSMO dataset. \textbf{\#AvgTokens(A)} and \textbf{\#AvgTokens(S)} denote the average number of tokens in articles and reference summaries respectively.}
\label{tab:msmo}
\end{table}

\paragraph{Evaluation Metrics}
In text summarization tasks, the evaluation of summary quality usually employs the \textbf{ROUGE} metric \citep{lin2004rouge}, which quantifies the degree of lexical correspondence between the produced sentences and the reference summaries. All the ROUGE scores in this paper refer to the F-1 ROUGE scores calculated by official script. In addition, \citet{zhu-etal-2018-msmo} introduce the metric of image precision (\textbf{IP}) to assess the quality of the output image, delineating the methodology as follows:
\begin{equation}
    \mathrm{IP}=\frac{\mid\left\{ref_{img}\right\} \cap\left\{rec_{img}\right\} \mid}{\left|\left\{rec_{img}\right\}\right|},
\end{equation}
where $ref_{img}$ and $rec_{img}$ denote the reference images and the recommended ones.

\paragraph{Implementation Details}
Our model utilizes the released checkpoint\footnote{\url{https://huggingface.co/Yale-LILY/brio-cnndm-uncased}} of a BART-like model, BRIO \citep{liu-etal-2022-brio}, to initialize corresponding parameters. And we take released CLIP model \citep{radford2021learning}\footnote{\url{https://huggingface.co/openai/clip-vit-base-patch32}} as the teacher model for image selection knowledge distillation. For the image processing, we employ the vision feature extractor of BLIP-2 \citep{li2023blip}\footnote{\url{https://github.com/salesforce/LAVIS/tree/main/projects/blip2}} to get visual features. The number of the learned queries is set to 32, resulting in an allocation of 33 token positions within the encoder for each image. And we set the upper limit of image number to 8. Noting that we concatenate multimodal tokens together as the input for two dual-modal encoders, the maximum number of textual and entity tokens is constrained by the encoder's maximum context length  as well as the length of the image sequence.

The train set of MSMO dataset is partitioned into 20 discrete subsets. Therefore, we employ a cumulative training strategy, wherein the model undergoes iterative training on each subset in succession. After training on each subset, the model's parameters are saved as checkpoints and evaluated on validation set. We identify the top-3 checkpoints as determined by the minimal validation loss. Subsequently, we compute and present the mean results derived from these checkpoints on test set.

In the process of image selection, we choose the image with greatest score as computed in Eq.(\ref{image-score}). And for text summarization, we use beam search with a beam size of 5 in decoding.

\paragraph{Baseline Models}
To demonstrate the efficacy of the proposed model, we conduct comparative analyses with extant methodologies in both text-based and multimodal summarization domains:
\begin{itemize}
    \item \textbf{BertSum} \citep{liu-lapata-2019-text} uses a general framework for both extractive and abstractive text summarization, with its encoder based on BERT \citep{kenton2019bert}. It has several raviants, out of which \textbf{BertAbs} and \textbf{BertExtAbs} can be used for abstractive text summarization.

    \item \textbf{BART} \citep{lewis-etal-2020-bart}, constructed as a denoising autoencoder, employs a sequence-to-sequence framework with significant applicability in the domain of text summarization.

    \item \textbf{ATG/ATL/HAN} utilizes a pointer-generator network \citep{see-etal-2017-get} and a multimodal attention mechanism, with variants reflecting different image representation approaches for attention operations.

    \item $\textbf{MOF}^{RR}$ \citep{zhu2020multimodal} ranks images via ROUGE score comparison of captions to textual reference, forming a visual reference. Variants of incorporating different hidden states into image discriminator are denoted as $\textbf{MOF}^{RR}_{enc}$ and $\textbf{MOF}^{RR}_{dec}$.

    \item \textbf{UniMS} \citep{zhang2022unims} proposes to merge textual and visual data to BART \citep{lewis-etal-2020-bart} encoder to construct a multimodal representation. Subsequently, it employs a visually guided decoder to integrate textual and visual modalities in guiding abstractive text generation.
\end{itemize}

\subsection{Experimental Result}
\begin{table}[]
    \centering
    \begin{tabular}{l|cccc}
        \toprule
        Model & R-1 & R-2 & R-L & IP \\
        \midrule
        \multicolumn{5}{c}{ Text Abstractive } \\
        \midrule
        BertAbs* & 39.02 & 18.17 & 33.20 & - \\
        BertExtAbs* & 39.88 & 18.77 & 38.36 & - \\
        BART & 42.93 & 19.95 & 39.97 & - \\
        \midrule
        \multicolumn{5}{c}{ Multimodal Abstractive } \\
        \midrule
        ATG* & 40.63 & 18.12 & 37.53 & 59.28 \\
        ATL* & 40.86 & 18.27 & 37.75 & 62.44 \\
        HAN* & 40.82 & 18.30 & 37.70 & 61.83 \\
        $\text{MOF}^{RR}_{enc}$* & 41.05 & 18.29 & 37.74 & 62.63 \\
        $\text{MOF}^{RR}_{dec}$* & 41.20 & 18.33 & 37.80 & 65.45 \\
        UniMS* & 42.94 & 20.50 & 40.96 & 69.38 \\
        \midrule
        EGMS & \textbf{44.47} & \textbf{21.20} & \textbf{41.43} & \textbf{75.81} \\
        \bottomrule
    \end{tabular}
    \caption{Experimental results for multimodal summarization on MSMO dataset. Results marked by * are taken from respective papers and \citet{zhang2022unims}.}
    \label{tab:exp-main}
\end{table}
As shown in Table \ref{tab:exp-main}, our proposed EGMS method outperforms all baselines in all metrics, which proves the effectiveness of our method and the necessity to incorporate knowledge graphs. 

The outcomes of this study reveal a number of intriguing phenomena:
(1) By fine-tuning BART for summarization task, it can achieve competitive results with models that introduce visual information. This proves that BART exhibits robust language modeling proficiencies, thereby indicating its substantial potential for applications in multimodal information modeling. The findings herein reinforce the rationale for its deployment in our modeling endeavors. (2) The UniMS framework, also based on BART model, has shown great improvements, especially in ROUGE-2 and ROUGE-L scores. This advancement suggests that the integration of visual data facilitates the model's capacity to process and interpret extended text sequences, surpassing the merely word-level analyses. Such findings are consistent with our initial hypothesis, which postulates that the incorporation of entity-level information rather than word-level would yield a more robust understanding of the multimodal data.

\subsection{Ablation Study}
\begin{table}[]
    \centering
    \begin{tabular}{l|cccc}
        \toprule
        Model & R-1 & R-2 & R-L & IP \\
        \midrule
        EGMS & \textbf{44.47} & \textbf{21.20} & \textbf{41.43} & \textbf{75.81} \\
        ~ -w/o IS & 44.25 & 21.05 & 41.21 & - \\
        ~ -w/o EI & 44.29 & 21.10 & 41.22 & 75.65 \\
        ~ -w/o TI & 44.35 & 21.07 & 41.35 & 62.88 \\
        \bottomrule
    \end{tabular}
    \caption{Ablation experiments on MSMO dataset. 'IS' stands for Image Selection module. 'EI' and 'TI' refer to the encoded visual representations derived from Entity-Image Encoder and Text-Image Encoder respectively.}
    \label{tab:exp-ablation}
\end{table}
In this subsection, we conduct ablation experiments to prove the effectiveness of different components of EGMS model. We remove Image Selection (IS) module, image representations derived from Entity-Image Encoder (EI) and Text-Image Encoder (TI) respectively. More specifically, by removing Image Selection module, we reduce MSMO problem to a multimodal summarization task with only textual output. Removing 'EI' means that we only use the encoded visual representations from Text-Image Encoder for summary generation and image selection. To elaborate, the weight $w_{te}$ from Eq.(\ref{weight}) is fixed to 1. Likewise, when removing 'TI', reliance is exclusively placed on the visual representations from Entity-Image Encoder, with the corresponding weight being constrained to 0.

The results are listed in Table \ref{tab:exp-ablation}. Analysis of the data reveals a consistent decline across all ablation variants, thereby demonstrating the validity and non-redundancy of our proposed EGMS method. Besides, we can find that the entity information predominantly enhances the capacity of the model to generate concise summaries, while the improvement of the model's image selection accuracy is smaller. This differential impact suggests that comprehensive textual data may suffice for the selection of pertinent images. However, the integration of additional entity information can have an advantage in the precise identification of salient components, aligning well with the core requirements of the summarization task.

\subsection{Parameter Sensitivity}

\begin{figure}[h]
\centering
\includegraphics[width=1.0\columnwidth]{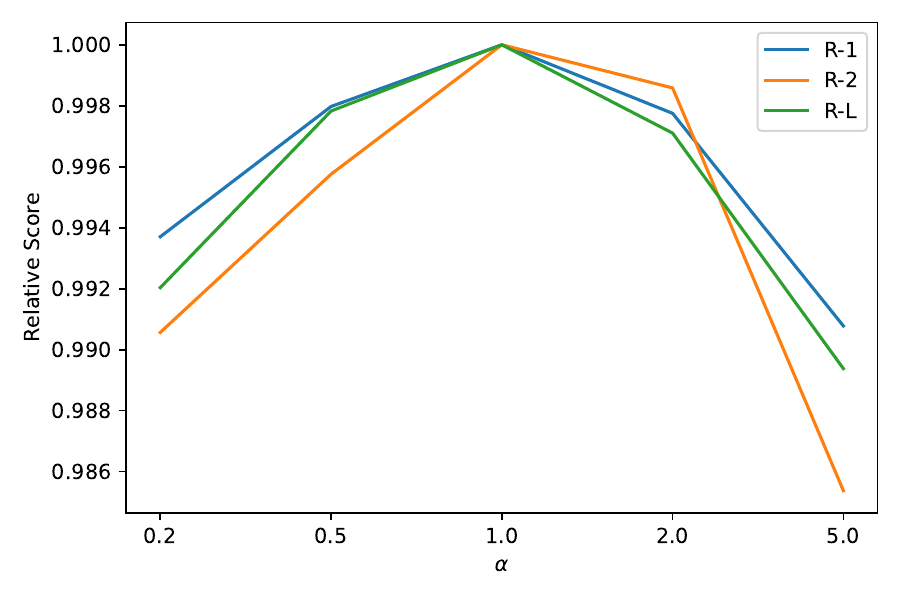}
\caption{Hyperparameter study on MSMO dataset. The results in the graph are normalized by the result of the corresponding metric with $\alpha = 1.0$.}
\label{fig:exp-hyper}
\end{figure}
To study the impact of the loss hyperparameter $\alpha$ in EGMS, a series of parameter sensitivity analyses were performed on the MSMO dataset. The results are reported in Figure \ref{fig:exp-hyper}. $\alpha = 1.0$ is the best hyperparameter of our model. From the results, we can see that larger or smaller $\alpha$ will lead to decrease on the summarization performance. This is reasonable as the hyperparameter controls the weight of the Image Selection loss in the total loss. A large weight will affect the Abstractive Summarization loss, while a small weight reduces the usefulness of the text-image multimodal knowledge aids learned from the teacher model in modeling multimodal information.

\subsection{Human Evaluation}
\begin{table}[]
    \centering
    \begin{tabular}{l|cccc}
        \toprule
        \multirow{2}*{Model} & \multicolumn{2}{c}{Text} & Image \\
         ~ & Coherence & Relevance & Relevance \\
        \midrule
        BART & 3.47 & 3.22 & - \\
        EGMS & \textbf{4.20} & \textbf{4.02} & \textbf{3.66} \\
        ~ -w/o IS & 3.75 & 3.64 & - \\
        ~ -w/o EI & 3.84 & 3.64 & 3.53 \\
        ~ -w/o TI & 3.84 & 3.67 & 3.45 \\
        \bottomrule
    \end{tabular}
    \caption{Human evaluation of different model outputs.}
    \label{tab:exp-human}
\end{table}
To further evaluate our models performance, we randomly select 120 data samples from test set for human evaluation. Subsequently, three graduate students are enlisted to evaluate them on a scale ranging from one to five, addressing various qualitative aspects. For abstractive text summarization, \textit{coherence} measures whether the summary is smooth and fluent. And \textit{relevance} assesses the extent to which the summary content corresponds with the information presented in the original document. For image selection, \textit{relevance} indicates the text-image relevance of the multimodal summary. Table \ref{tab:exp-human} indicates that our method can generate more coherent and relevant summaries compared to other variants and baselines.

\section{Conclusions}
In this paper, we propose an Entity-Guided Multimodal Summarization model (EGMS), that incorporates entity-specific information into solving MSMO problem. Based on BART, our model introduces a pair of multimodal encoders with shared weights to concurrently process text-image and entity-image information. Subsequently, a gating mechanism is used to fuse the visual representations, which will further be utilized in the generation of textual summaries. As for image selection, we also use a gating mechanism and distill knowledge from a pre-trained vision-language model. Extensive experiments on public MSMO dataset demonstrat the effectiveness of our proposed method. We hope our work could lead to more future studies in this field.

\section{Limitations}
In our proposed EGMS method, incorporating the knowledge graph requires the entity recognition process, which will consume additional time compared with other MSMO methods. And if we need to use other domains' knowledge graphs, it will be requisite to undertake retraining of the entity representations and the model. However, by utilizing a general-purpose knowledge graph, our model can be applied in most scenarios.
Another limitation is that since the MSMO dataset is labeled with pictorial references only on the test set, we adopt a method that utilizes knowledge distillation for image selection learning. And the results of such an approach can be affected by the performance of the teacher vision-language model.


\section*{Acknowledgements}

This research was partial supported by grants from the National Key Research and Development Program of China (Grant No. 2021YFF0901000), the National Natural Science Foundation of China (No. U20A20229), Anhui Provincial Natural Science Foundation (No. 2308085QF229), and the Fundamental Research Funds for the Central Universities (No. WK2150110034).

\bibliography{arxiv}
\bibliographystyle{acl_natbib}

\appendix

\section{Multimodal Summary Sample}

\begin{figure}[h]
\centering
\includegraphics[width=1.0\columnwidth]{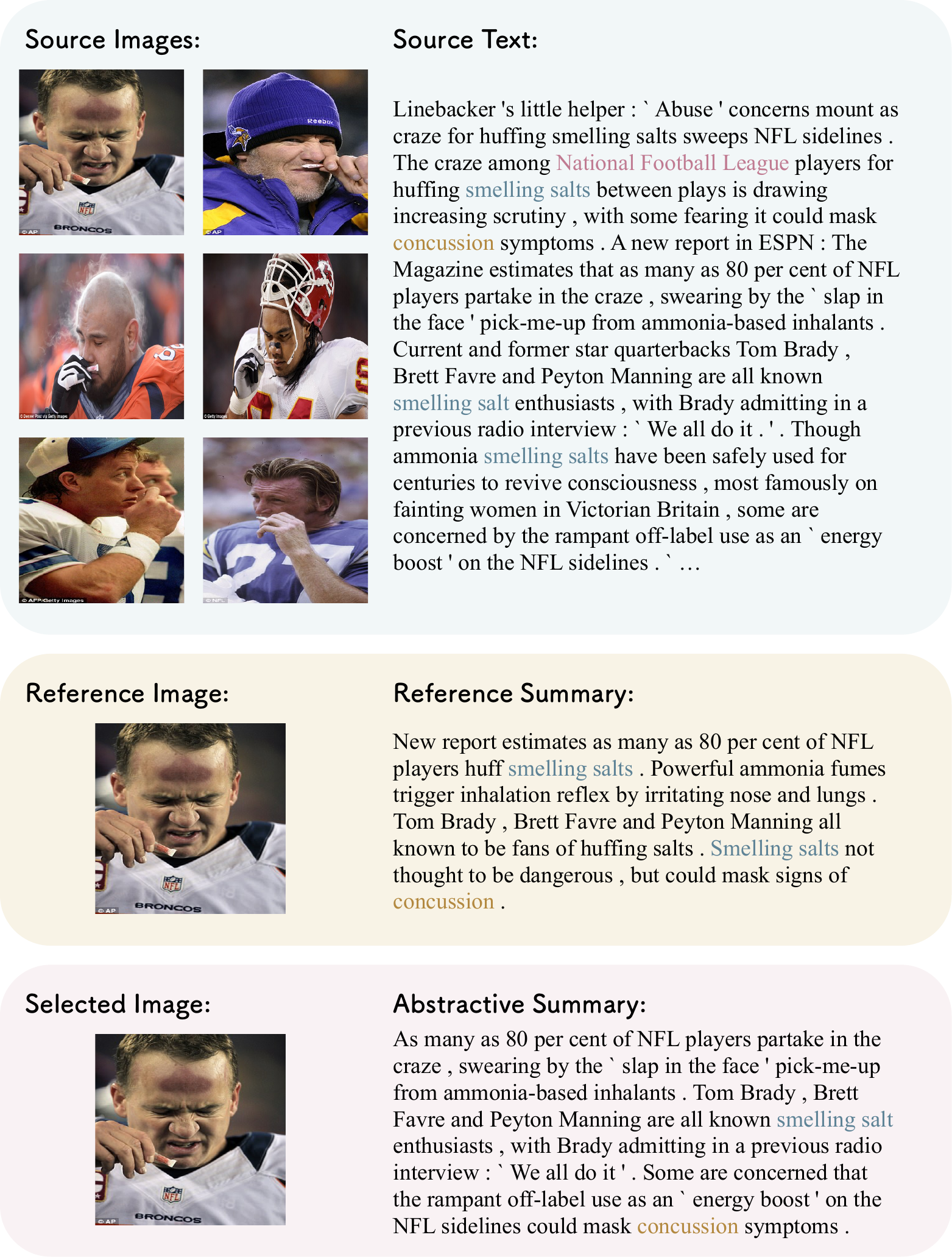}
\caption{An example of multimodal summary.}
\label{fig:case-appendix}
\end{figure}

To better show the effectiveness of our proposed EGMS method, we illustrate a case study in Figure~\ref{fig:case-appendix}.
From this figure, we can find that our model accurately recognizes the entity \textit{smelling salts}. And each image in the source input contains information about it. When considering a word-level approach, the isolated word \textit{salts} is not able to get the corresponding meaning accurately. However, the incorporation of entity-level information allows for an enhanced understanding of the correlations between textual data and visual elements, thereby improving the model's capacity for multimodal learning.

\section{Entity Analysis}

\begin{figure}[ht]
\centering
\includegraphics[width=1.0\columnwidth]{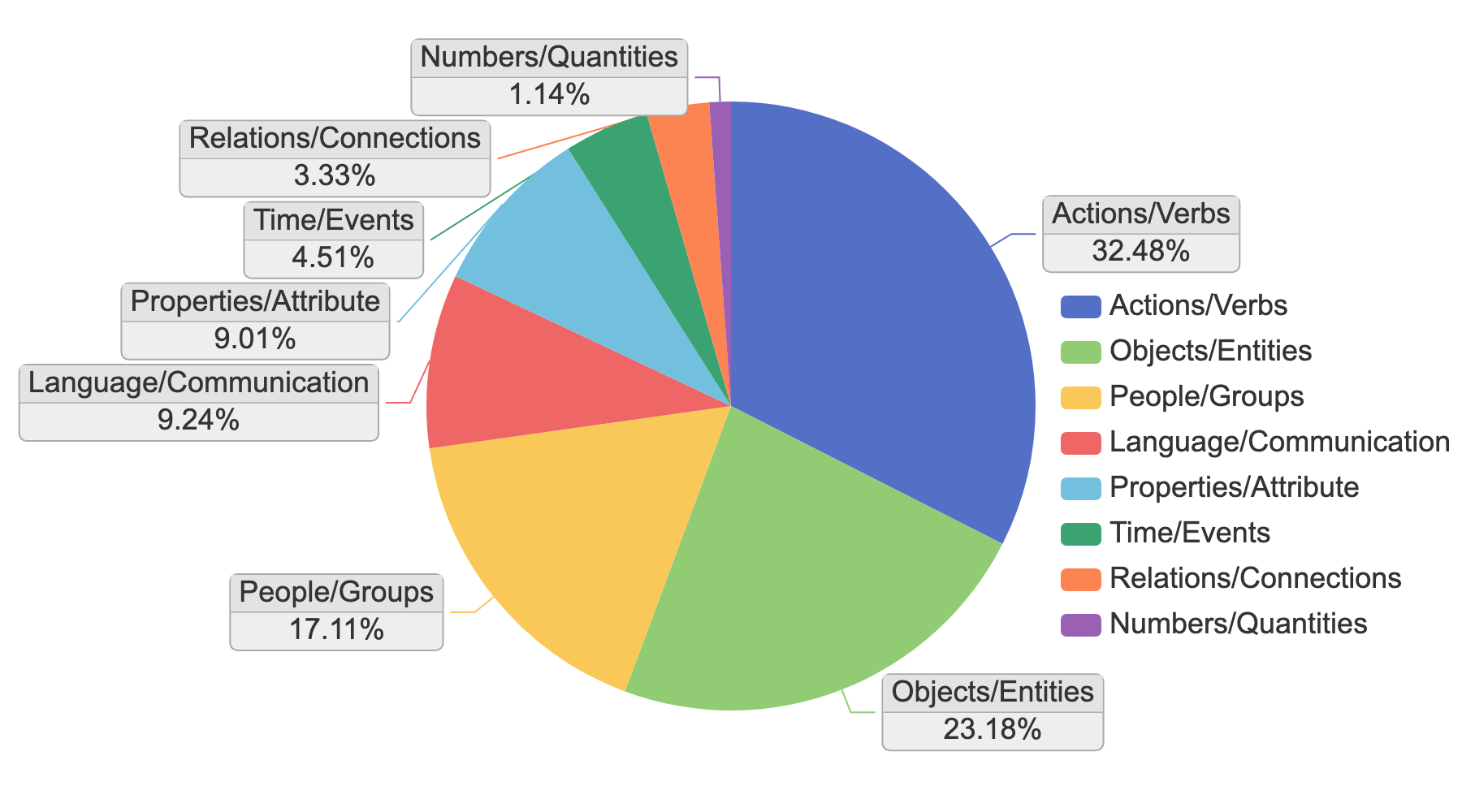}
\caption{Quantitative analysis of entities.}
\label{fig:entity}
\end{figure}

We employ the widely used ConceptNet\footnote{\url{https://conceptnet.io}} as our base knowledge graph. However, ConceptNet lacks labeling information for entity categories, necessitating manual labeling. We first set eight categories for entities within ConceptNet as follows:

\begin{itemize}
    \item \textbf{Actions/Verbs}: Actions (check in); Activities (pole vault)
    \item \textbf{Objects/Entities}: Physical Objects (coffee table); Abstractive Concepts (global warming)  
    \item \textbf{People/Groups}: Individuals (Michael Jackson); Organizations (Apple Inc.)
    \item \textbf{Language/Communication}: Linguistic Units (words or phrases or sentences); Languages (Standard Chinese)
    \item \textbf{Properties/Attribute}: Physical Properties (red); Abstractive Properties (warm hearted)
    \item \textbf{Time/Events}: Events (World War II); Time Units (year, hour)
    \item \textbf{Relations/Connections}: Relationships (part of); Locations (on top)
    \item \textbf{Numbers/Quantities}: Quantities (light year); Numbers (1, 10)
\end{itemize}

We randomly selected several data instances and tasked several annotators with categorizing 3,063 entities extracted from these instances. The results, displayed in Figure~\ref{fig:entity}, indicate that most entities pertain to activities, objects, people, and organizations. These entities possess substantial practical meanings that can aid in comprehending the main content of the text.

\begin{table*}[]
    \centering
    \begin{tabular}{c|p{0.7\columnwidth}|c|p{0.7\columnwidth}}

        \toprule
        Variables & Meaning & Variables & Meaning \\
        \midrule
        $D$ & multimodal input & $T$ & textual input, article text \\
        $t_i$ & i-th token of the tokenized text & $P$ & visual input, in-article images \\
        $p_i$ & i-th image & $K$ & extracted entity list \\
        $k_i$ & i-th extracted entity & $S$ & multimodal summary \\
        $S_t$ & textual summary & $s_i$ & i-th token of the textual summary \\
        $S_i$ & pictorial summary, extracted subset of $P$ & $r_i$ & image features from frozen image encoder \\
        $v_i$ & queried image features & $e_t$ & textual embeddings \\
        $e_v$ & visual embeddings & $e_e$ & entity embeddings \\
        $e_{intra-pos}$ & position embeddings within single image & $e_{multi-pos}$ & multimodal position embeddings \\
        $h_{ti}$ & hidden states of the output layer of text-image encoder & $h_{ei}$ & hidden states of the output layer of entity-image encoder \\
        $h_{T_{ti}}$ & textual part of $h_{ti}$ & $h_{E_{ei}}$ & entity part of $h_{ei}$ \\
        $h_{V_{ti}}$ & visual part of $h_{ti}$ & $h_{V_{ei}}$ & visual part of $h_{ei}$ \\
        $h_{V_{ti-cls}}$ & output hidden states of $v_{CLS}$ from text-image encoder & $h_{V_{ei-cls}}$ & output hidden states of $v_{CLS}$ from entity-image encoder \\
        $h_{V_{comb}}$ & combined visual hidden states from two encoders & $h_{enc-hid}$ & combined hidden states of two encoders \\
        \bottomrule
    \end{tabular}
    \caption{Table for notations.}
    \label{tab:notation}
\end{table*}

\end{document}